\documentclass{article}

\usepackage[preprint,nonatbib]{neurips_2020}
\usepackage[sort&compress,numbers]{natbib}

\usepackage[utf8]{inputenc} 
\usepackage[T1]{fontenc}    
\usepackage{hyperref}       
\usepackage{url}            
\usepackage{booktabs}       
\usepackage{amsfonts}       
\usepackage{amsmath}
\usepackage{graphicx}
\usepackage{subcaption}
\usepackage{nicefrac}       
\usepackage{microtype}      

\title{Auto-Encoding for Shared Cross Domain Feature Representation and Image-to-Image Translation}

\author{%
  Safalya Pal \\
  Department of Statistics\\
  Amity University\\
  Kolkata, West Bengal, India \\
  \texttt{safalya.pal@student.amity.edu} \\
}

\begin{document}

\maketitle

\begin{abstract}
Image-to-image translation is a subset of computer vision and pattern recognition problems where our goal is to learn a mapping between input images of domain $\mathbf{X}_1$ and output images of domain $\mathbf{X}_2$. Current methods use neural networks with an encoder-decoder structure to learn a mapping $G:\mathbf{X}_1 \to\mathbf{X}_2$ such that the distribution of images from $\mathbf{X}_2$ and $G(\mathbf{X}_1)$ are identical, where $G(\mathbf{X}_1) = d_G (f_G (\mathbf{X}_1))$ and $f_G (\cdot)$ is referred as the encoder and $d_G(\cdot)$ is referred to as the decoder. Currently, such methods which also compute an inverse mapping $F:\mathbf{X}_2 \to \mathbf{X}_1$ use a separate encoder-decoder pair $d_F (f_F (\mathbf{X}_2))$ or at least a separate decoder $d_F (\cdot)$ to do so. Here we introduce a method to perform cross domain image-to-image translation across multiple domains using a single encoder-decoder architecture. We use an auto-encoder network which given an input image $\mathbf{X}_1$, first computes a latent domain encoding $Z_d = f_d (\mathbf{X}_1)$ and a latent content encoding $Z_c  = f_c (\mathbf{X}_1)$, where the domain encoding $Z_d$ and content encoding $Z_c$ are independent. And then a decoder network $g(Z_d,Z_c)$ creates a reconstruction of the original image $\mathbf{\widehat{X}}_1=g(Z_d,Z_c )\approx \mathbf{X}_1$. Ideally, the domain encoding $Z_d$ contains no information regarding the content of the image and the content encoding $Z_c$ contains no information regarding the domain of the image. We use this property of the encodings to find the mapping across domains $G: X\to Y$ by simply changing the domain encoding $Z_d$ of the decoder's input. $G(\mathbf{X}_1 )=d(f_d (\mathbf{x}_2^i ),f_c (\mathbf{X}_1))$ where $\mathbf{x}_2^i$ is the $i^{th}$ observation of $\mathbf{X}_2$.
\end{abstract}

\section{Introduction}

Humans have always been able to percieve the similarity in structures of objects which vary vastly in nature. If we're given an illustration and a photograph of the same person, our minds easily spot the similarity in their scemantic features. 

Recent developments in generative models has greatly improved the quality of algorithms which can create generalized feature representations\citep{Jerry18, Royer18}, in which semantically similar objects across varying domains are placed closely in the encoding space and dissimilar ones are placed far apart. 

In this work, we try to learn separate encodings which help us discriminate between domains but also find a semantic similarity among objects across those domains. Recent models have performed good in mapping encodings which stay semantically consistent across domains but none have learned to create encodings which are domain specific.

Perhaps the most similar recent example would be XGAN\citep{Royer18}. Which, given images from two domains $\mathbf{X}_1$ and $\mathbf{X}_2$, uses an encoder $f_1$ and a decoder $g_1$ for domain $\mathbf{X}_1$ and an another encoder $f_2$ and decoder $g_2$ for domain $\mathbf{X}_2$ to learn a mappings $G_1: \mathbf{X}_1 \to \mathbf{X}_2$ and $G_2 : \mathbf{X}_2 \to \mathbf{X}_1$ by enforcing a cross-domain consistency between the encoders' output encodings. 

\section{Related Work}

\paragraph{Domain Transfer}
DTN\citep{Yaniv16} transfers images from a source domain to the target domain while keeping their semantic features similar. It contains a pretrained feature extractor $f$ and a generator $g$ on top of the output of $f$. the DTN is trained using an adversarial loss to keep the outputs believable and a feature consistency loss to preserve the semantic features across the domains. 

\paragraph{Image-to-Image Translation}
The image translation network most related to our approach would be XGAN\citep{Royer18} , which uses dual auto-encoders $g_1 \circ f_1$ and $g_2 \circ f_2$ on domains $\mathbf{X}_1$ and $\mathbf{X}_2$. It encourages the encodings of the encoders $f_1$ and $f_2$ to lie in the same subspace, i.e., it encourages the encodings to be indistinguishable. For this, it trains a binary classifier $q$ on top of the latent encodings to categorize the images as coming from either $\mathbf{X}_1$ or $\mathbf{X}_2$. $q$ is trained to maximize the classification accuracy while the encoders $f_1$ and $f_2$ similtenously learn to decrease it, i.e. to confuse the classifier. It also enforces the encodings to preserve the semantic feature after image translation by using a semantic consistency loss\citep{Jun17} between the original image's latent encoding and the translated image's feature encoding. 

\begin{figure}[h]
\centering
\begin{subfigure}{1\textwidth}
  \centering
  \includegraphics[width = 0.6\linewidth]{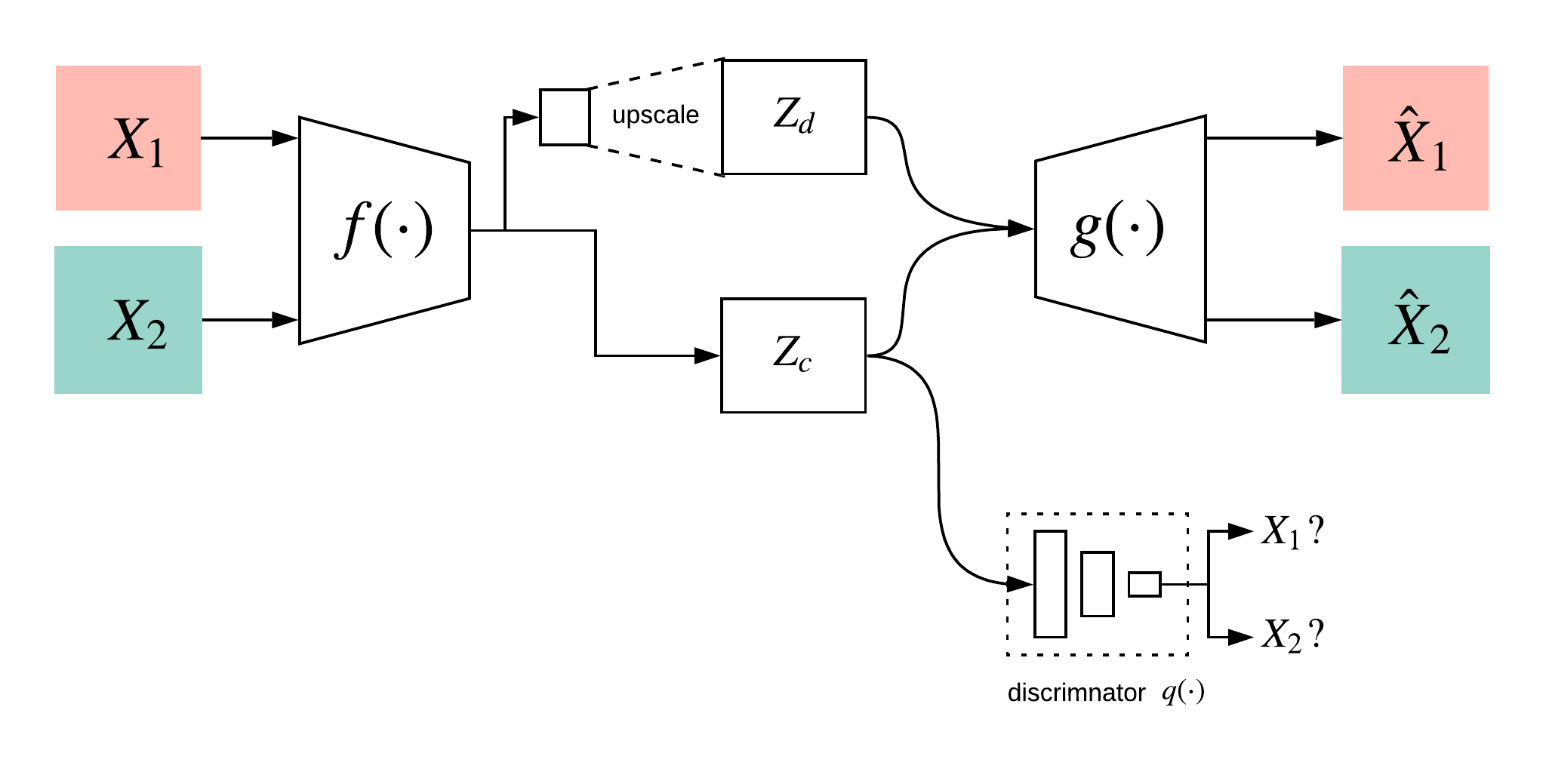}
  \caption{SRAE architecture with a single discriminator $q(Z_c)$}
  \label{fig:arc1}
\end{subfigure}
\begin{subfigure}{1\textwidth}
  \centering
  \includegraphics[width = 0.6\linewidth]{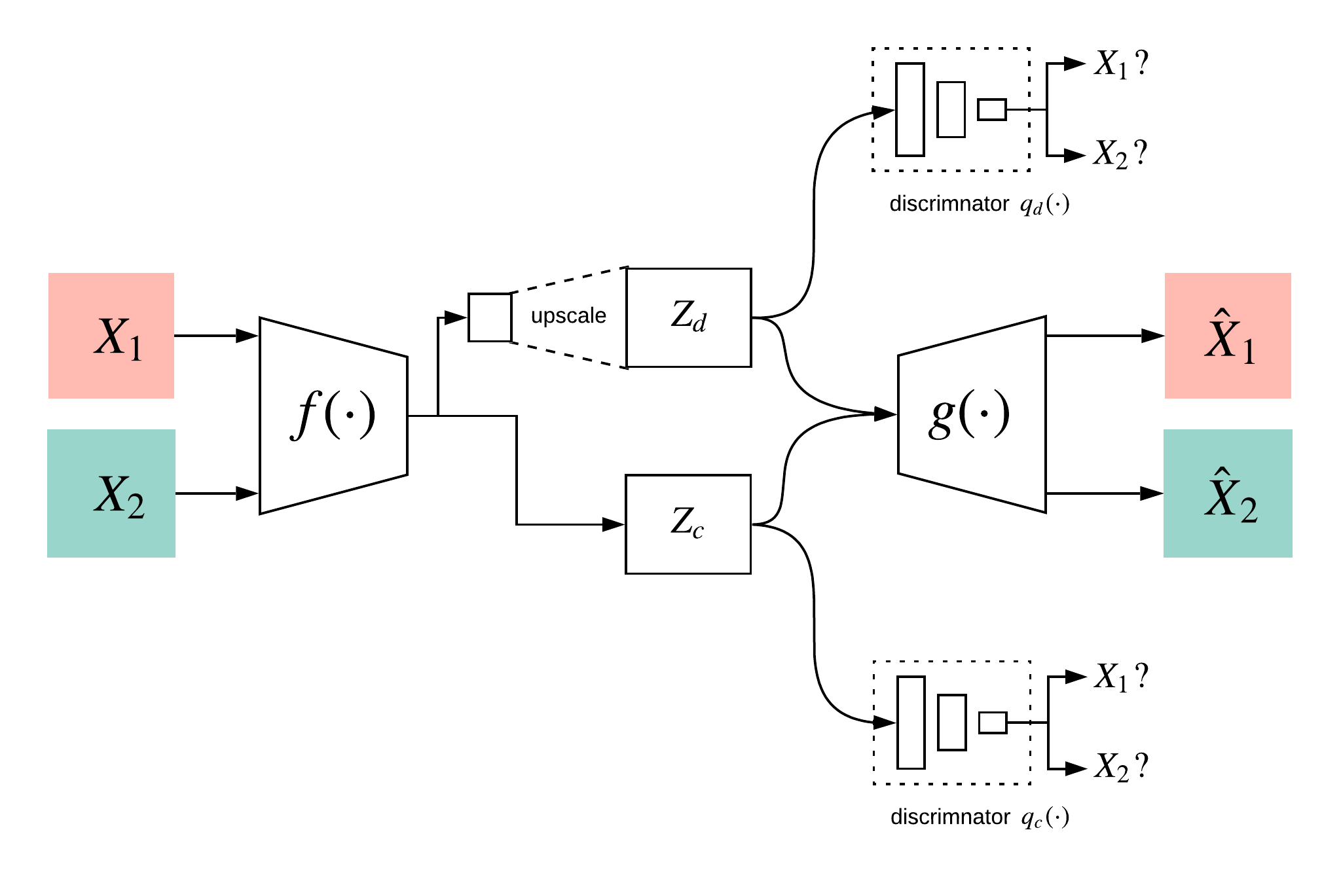}
  \caption{SRAE architecture with two discriminators $q_c(Z_c)$ and $q_d(Z_d)$}
  \label{fig:arc2}
\end{subfigure}
\caption{diagrams of SRAE architectures with one discriminator (\subref{fig:arc1}) and two discriminators (\subref{fig:arc2})}
\label{fig:architecture}
\end{figure}

\section{Method}
\subsection{Objective}Given unpaired samples from two domains $\mathbf{X}_1$ and $\mathbf{X}_2$ having semantically similar objects, we want an \textbf{encoder} $f(X)$ to learn two separate latent encodings $Z_d \in  \mathbb{R}^{ a \times b \times j }$ and $Z_c \in  \mathbb{R}^{z \times b \times k}$, were $Z_d$ is called the \textbf{"domain encoding"}, which only gives us information regarding the input image's domain and $Z_c$ is called the \textbf{"content encoding"}, which only gives us information regarding the semantic contents of the input image. We also want a \textbf{decoder} $g(Z_d, Z_c)$ to create a reconstruction $\widehat{\mathbf{X}}$ of the original input $\mathbf{X}$ from the latent encodings. We introduce an architecture called \textbf{Split Representation Auto-Encoder (SRAE)} with two variations, one which uses a single \textbf{discriminator} $q(Z_c)$ to predict the domain of the input image from the latent content encoding $Z_c$ and the other uses two \textbf{discriminators} $q_c(Z_c)$ and $q_d(Z_d)$ to predict the domain of the input image from the latent content encoding $Z_c$ and the latent domain encoding $Z_d$ respectively.


\subsection{Architecture}
\begin{figure}[h]
\centering
\begin{subfigure}{1\textwidth}
  \centering
  \includegraphics[scale = 0.125]{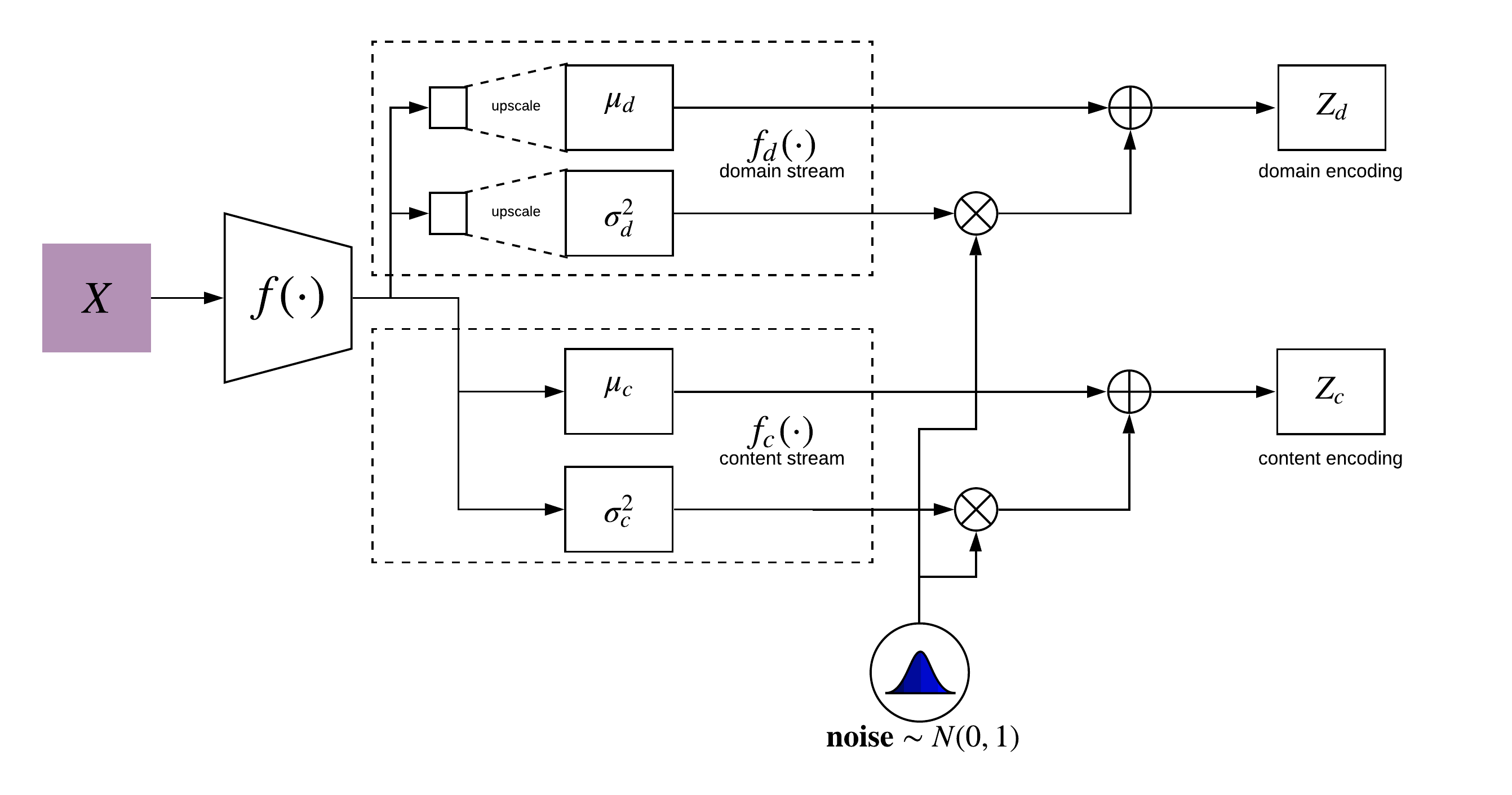}
  \caption{encoder}
  \label{fig:sub1}
\end{subfigure}
\newline
\begin{subfigure}{1\textwidth}
  \centering
  \includegraphics[scale = 0.125]{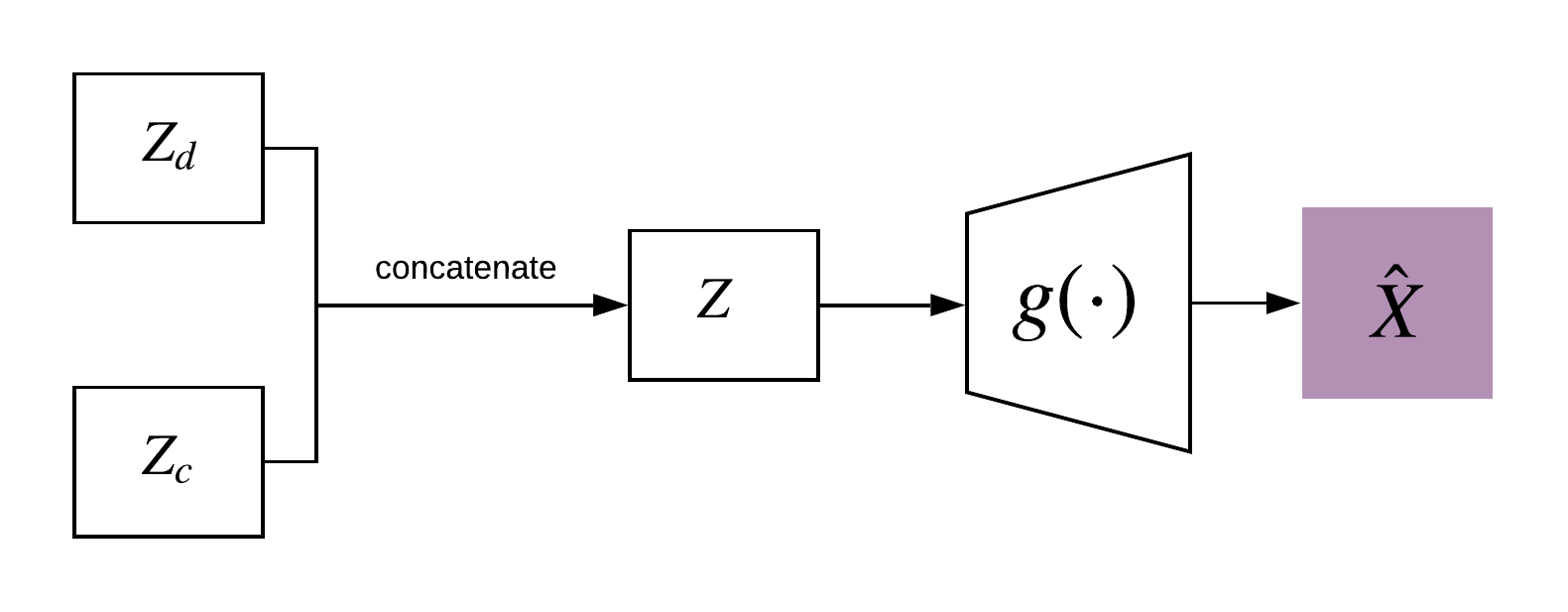}
  \caption{decoder}
  \label{fig:sub2}
\end{subfigure}
\caption{diagram of the encoder (\subref{fig:sub1}) and decoder (\subref{fig:sub2})}
\label{fig:test}
\end{figure}

\paragraph{Encoder}
The first part of our SRAE's \textbf{encoder} (figure \ref{fig:sub1}) takes in an input image $\mathbf{X}$ and passes it through convolutional layers. We denote this initial part as $f_\phi$ and with parameters $\Theta_\phi$. The next part splits the encoder into two separate streams, $f_c$ (\textbf{content stream}) and $f_d$ (\textbf{domain stream}), which are parameterized by $\Theta_c$ and $\Theta_d$ respectively. Both the streams output parameters of a normal distribution. The \textbf{content stream}'s outputs, $\mu_c$ and $\sigma^2_c$ are both $a \times  b \times  k$ dimentional, i.e. $\mu_c, \sigma^2_c \in \mathbb{R}^{a \times b \times k}$. While the \textbf{domain stream}'s outputs are both $1 \times  1 \times  j$ dimentional and are then upscaled to produce $\mu_d, \sigma^2_d \in \mathbb{R}^{a \times b \times j}$. We then sample $Z_c$ from the distribution $\mathcal{N}(\mu_c, \sigma^2_c )$ and $Z_d$ from the distribution $\mathcal{N}(\mu_d, \sigma^2_d )$. But since a sampling operation is non-differentiable, we use the reparameterization trick introduced in \textbf{VAEs}\citep{Died14} and sample a \textbf{noise} $\mathbf{\epsilon} \sim  \mathcal{N}(0, 1)$ and define the encodings as $$Z_c = \mu_c + \mathbf{\epsilon} \cdot \sigma^2_c$$ and $$Z_d = \mu_d + \mathbf{\epsilon} \cdot \sigma^2_d$$

\paragraph{Decoder}
We take the encodings ($Z_c$, $Z_d$) produced by our \textbf{encoder} and concatenate them to get $Z \in \mathbb{R}^{a \times b \times k + j}$. Our decoder $g$ takes $Z$ as an input to produce an image $\mathbf{\widehat{X}}$ such that $\mathbf{\widehat{X}} = g(Z) \approx \mathbf{X}$

\subsection{Learning}
\paragraph{Perceptual Loss}
Perceptual loss\citep{Hou17, John16} between two images is defined as the difference between the hidden features in a pretrained perceptual loss network $\mathcal{P}$, we use VGG-16 as the perceptual loss network over here. We denote the $i^{th}$ layer of $\mathcal{P}$ with image $\mathbf{X}$ as the input by ${\mathcal{P}}( \mathbf{X} )^i $. The \textbf{perceptual loss} is written as the following : $$\mathcal{L}_r = \sum_{i=1}^n \left \| {\mathcal{P}}(\mathbf{X})^i - \, {\mathcal{P}}(\mathbf{\widehat{X}})^i \right \| ^2 $$

\begin{figure}[h]
\centering
\includegraphics[width = 0.6\linewidth]{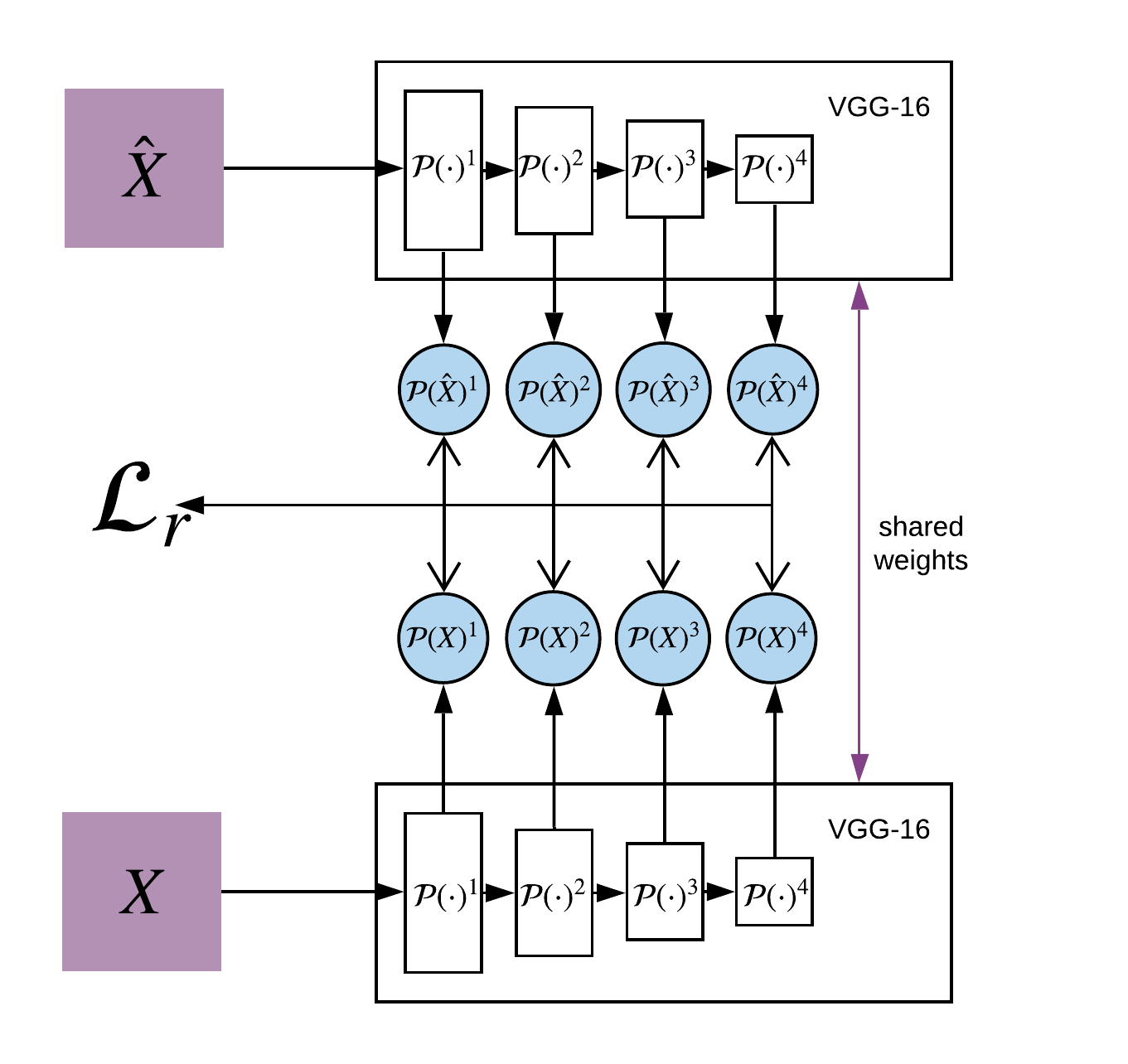}
\caption{calculation of perceptual loss}
\label{fig:per}
\end{figure}

\paragraph{Discriminator Loss}
Given an input image $\widehat{\mathbf{X}}$ which belongs to the domain $\mathbf{Y}$ (out of $m$ domains) and the corresponding latent encodings $Z_c$ and $Z_d$, for the first variation of our SRAE architecture with a single \textbf{discriminator} (\ref{fig:arc1}), the discriminator's loss is the cross-entropy between it's output probability distribution $q(Z_c)$ and the actual probability distribution defined as $$\mathcal{L}^c_q = \sum_{i=1}^m P(\mathbf{y}^i)\cdot log\left(q_c(Z_c)^i\right)$$ and for the second variation with two \textbf{discriminators} (\ref{fig:arc2}), the discriminator loss is defined as follows:  $$\mathcal{L}^d_q = \sum_{i=1}^m P(\mathbf{y}^i) \cdot log\left(q_d(Z_d)^i\right)$$ $$\mathcal{L}_q = \mathcal{L}^c_q + \mathcal{L}^d_q$$

\paragraph{Content Stream Loss}Since our objective is to make the \textbf{content encoding} $Z_c$ not contain any information about the domain $\mathbf{Y}$, a discriminator trained on the \textbf{content encoding} should not be able to predict the domain of the input image. Since the discriminator outputs a probability distribution of the categories, to achieve this, we maximize the \textbf{entropy} of the discriminator's outputs with respect to the \textbf{content stream}'s parameters $\Theta_c$. This loss is written as
$$\mathcal{L}_c = \mathbb{E}\left[ q_c(Z_c) \cdot log\left(q_c\left( Z_c \right) \right)\right]$$ $$= \frac{1}{m} \sum_{i=1}^m q_c(Z_c)^i \cdot log (q_c(Z_c)^i)$$
We use gradient ascent to update the \textbf{content stream} parameters $\Theta_c$ $$ \Theta_c := \Theta_c + \alpha_1 \nabla_{\Theta_c}{\mathcal{L}_c}$$

\paragraph{Domain Stream Loss}\footnote{Only in the variation of SRAE with two discriminators} To explicitly make the domain encoding $Z_d$ only contain information about the domain, we take the cross-entropy loss $\mathcal{L}^d_q$ between the discriminator's output $q_d(Z_d)$ and the labels and minimize it with respect to the \textbf{domain stream}'s parameters $\Theta_d$ $$ \Theta_d := \Theta_d - \alpha_2 \nabla_{\Theta_d}{\mathcal{L}_d}$$

\section{Experiments}
\subsection{Image Translation}
To perform Image-to-Image translation\citep{Jun17, Phil17, Anokhin_2020_CVPR} from domain $\mathbf{X}_1$ to $\mathbf{X}_2$, i.e. $G : \mathbf{X}_1 \to \mathbf{X}_2$, we take the \textbf{content encoding} $Z_c^1$ of our image $\mathbf{X}_1$ and the \textbf{domain encoding} $Z_d^2$ of any arbitrary image $\mathbf{x}_2^i \in \mathbf{X}_2$, where $i$ denotes the $i^{th}$ observation. We then pass the combined encodings through the \textbf{decoder} $g(Z_c, Z_d) $and define the mapping as $$G(X_1) = g(Z_c^1, Z_d^2)$$

\begin{figure}
\centering
\begin{subfigure}{0.5\textwidth}
  \centering
  \includegraphics[width = 0.8\linewidth]{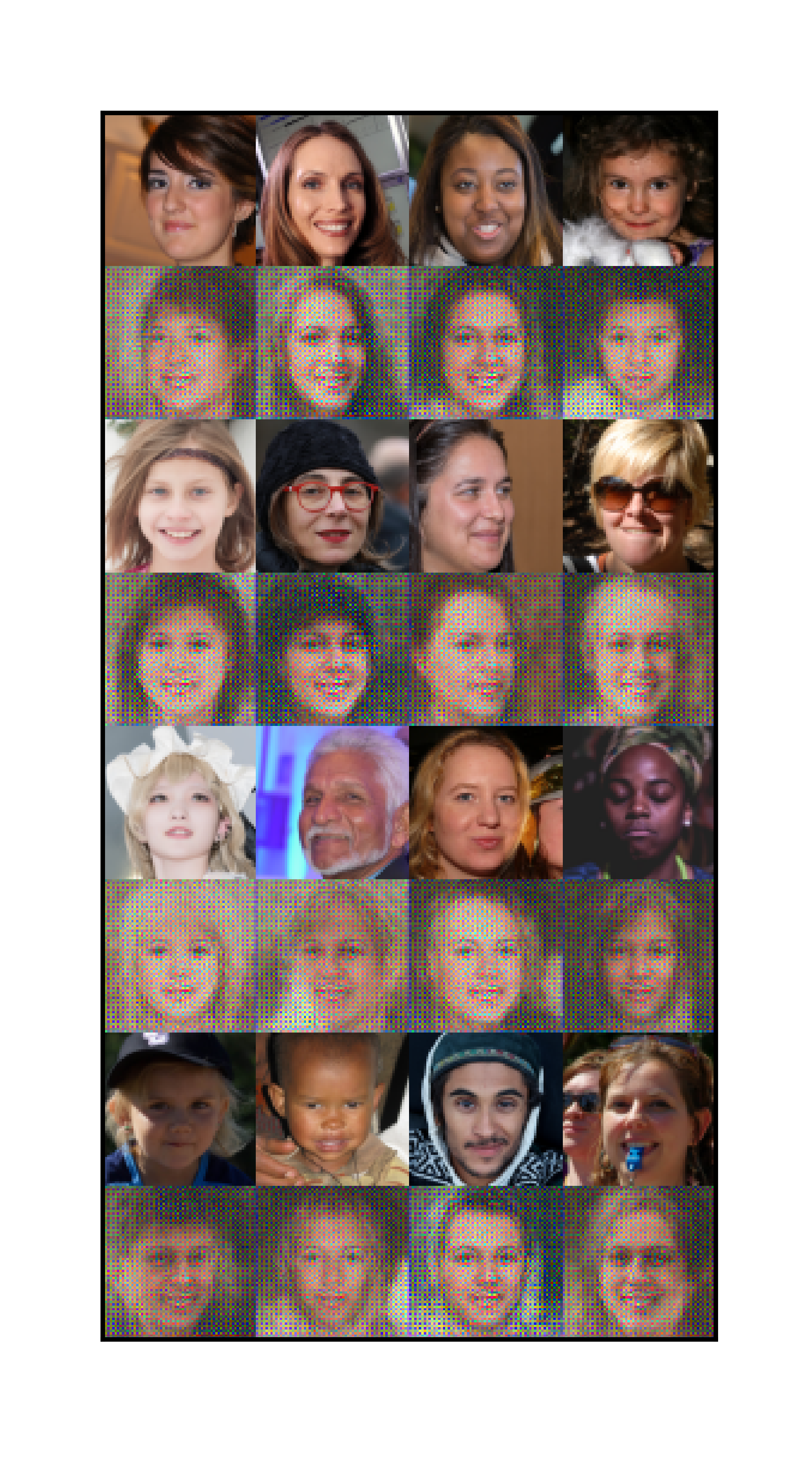}
  \caption{human faces}
  \label{fig:rec1}
\end{subfigure}%
\begin{subfigure}{0.5\textwidth}
  \centering
  \includegraphics[width = 0.8\linewidth]{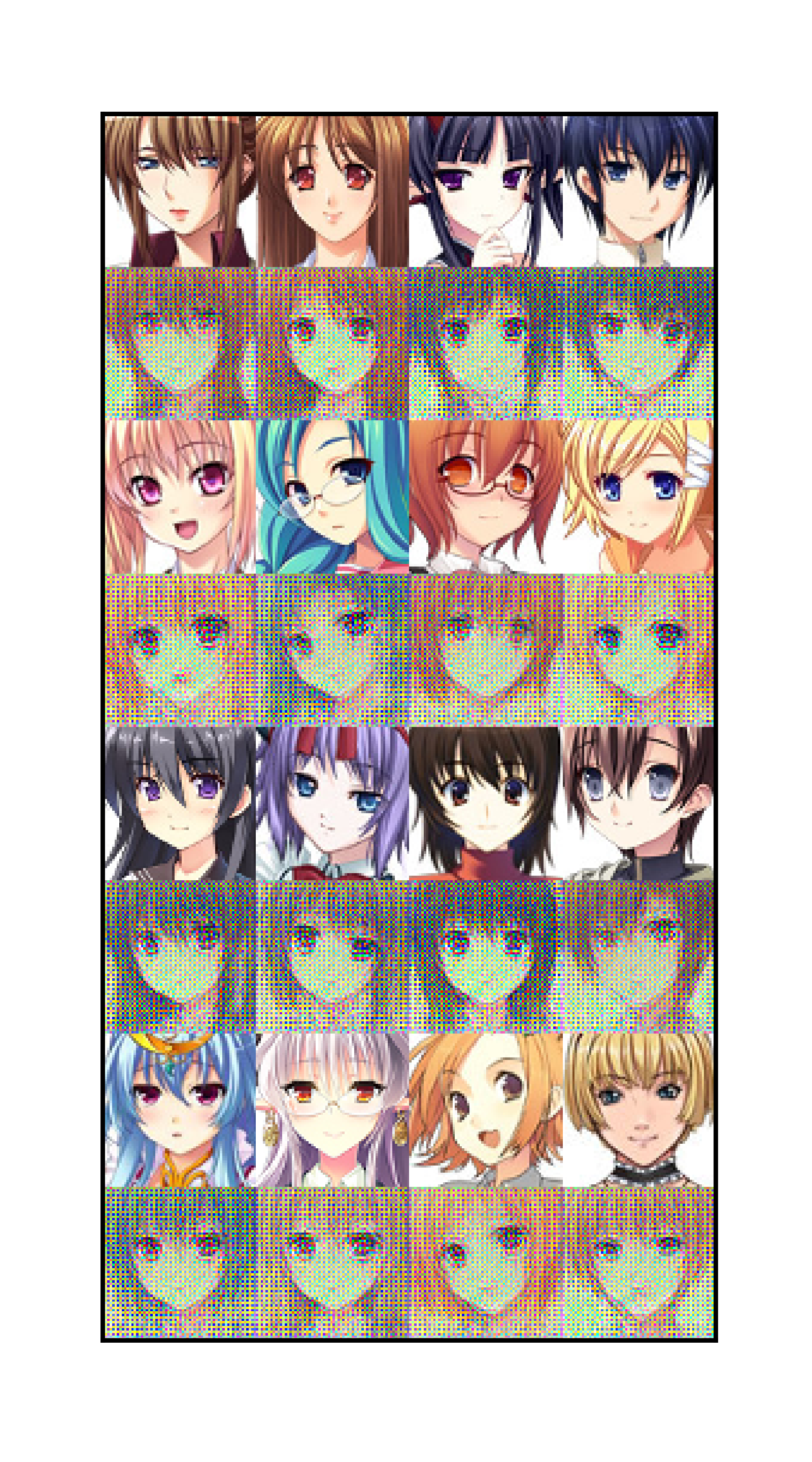}
  \caption{anime faces}
  \label{fig:rec2}
\end{subfigure}
\caption{reconstructions of images for human faces (domain \subref{fig:rec1}) and anime faces (domain \subref{fig:rec2}) \newline odd rows : ground truth ; even rows : reconstructed image}
\label{fig:imtran}
\end{figure}

For the first task (figure \ref{fig:imtran}), our network was trained on two datasets: FFHQ dataset\citep{Karras19}, originally consisting of 70,000 images but we used a smaller subset of 22,000 images scaled down to $64 \times 64$ pixels and 20,600 images from the Getchu anime face dataset\citep{getchu}	 scaled down to $64 \times 64$ pixels. 

We trained this network using the variation of SRAE with a single discriminator and no \textbf{domain stream loss}

\begin{figure}
\centering
\begin{subfigure}{0.5\textwidth}
  \centering
  \includegraphics[width = 0.8\linewidth]{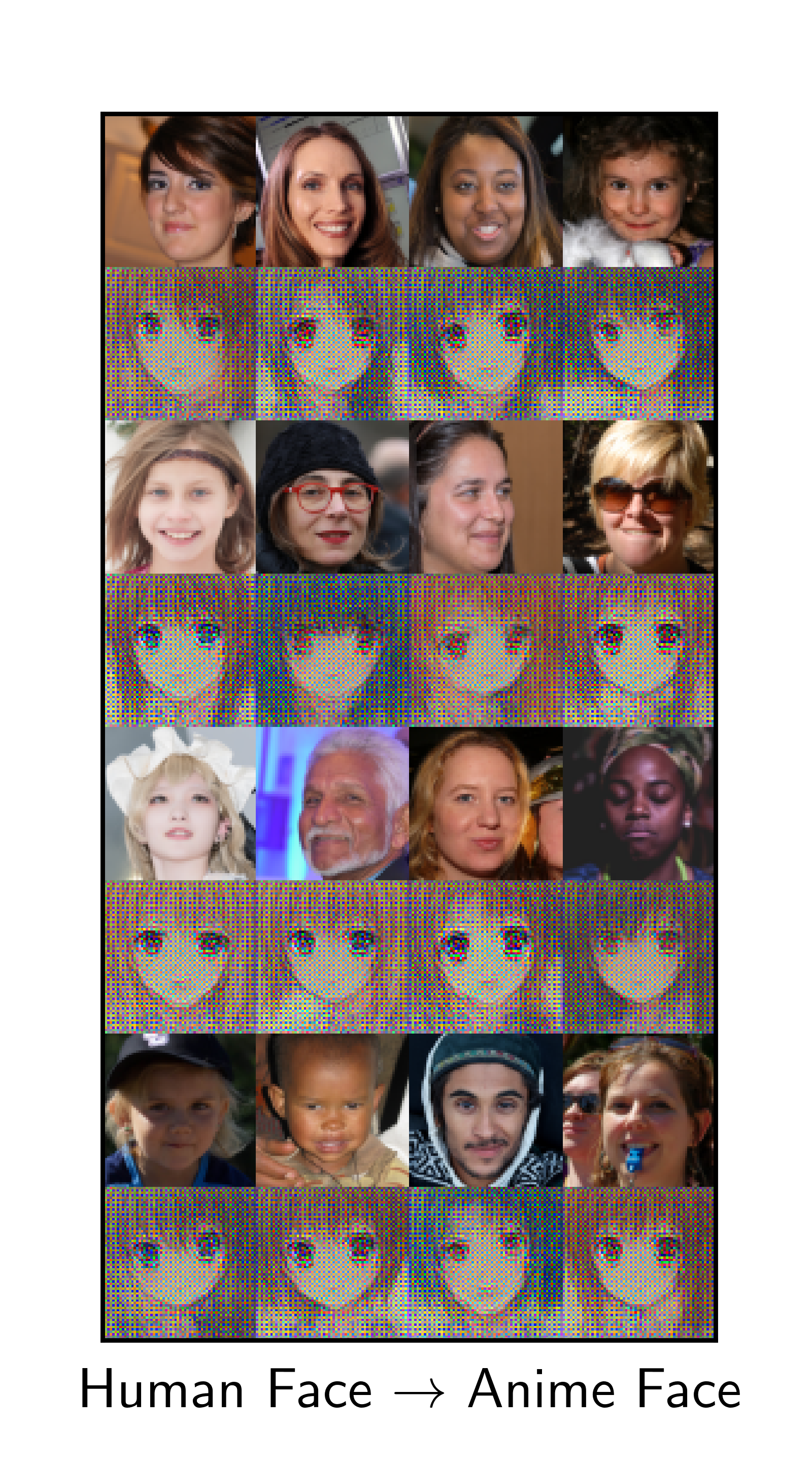}
  \caption{}
  \label{fig:ab}
\end{subfigure}%
\begin{subfigure}{0.5\textwidth}
  \centering
  \includegraphics[width = 0.8\linewidth]{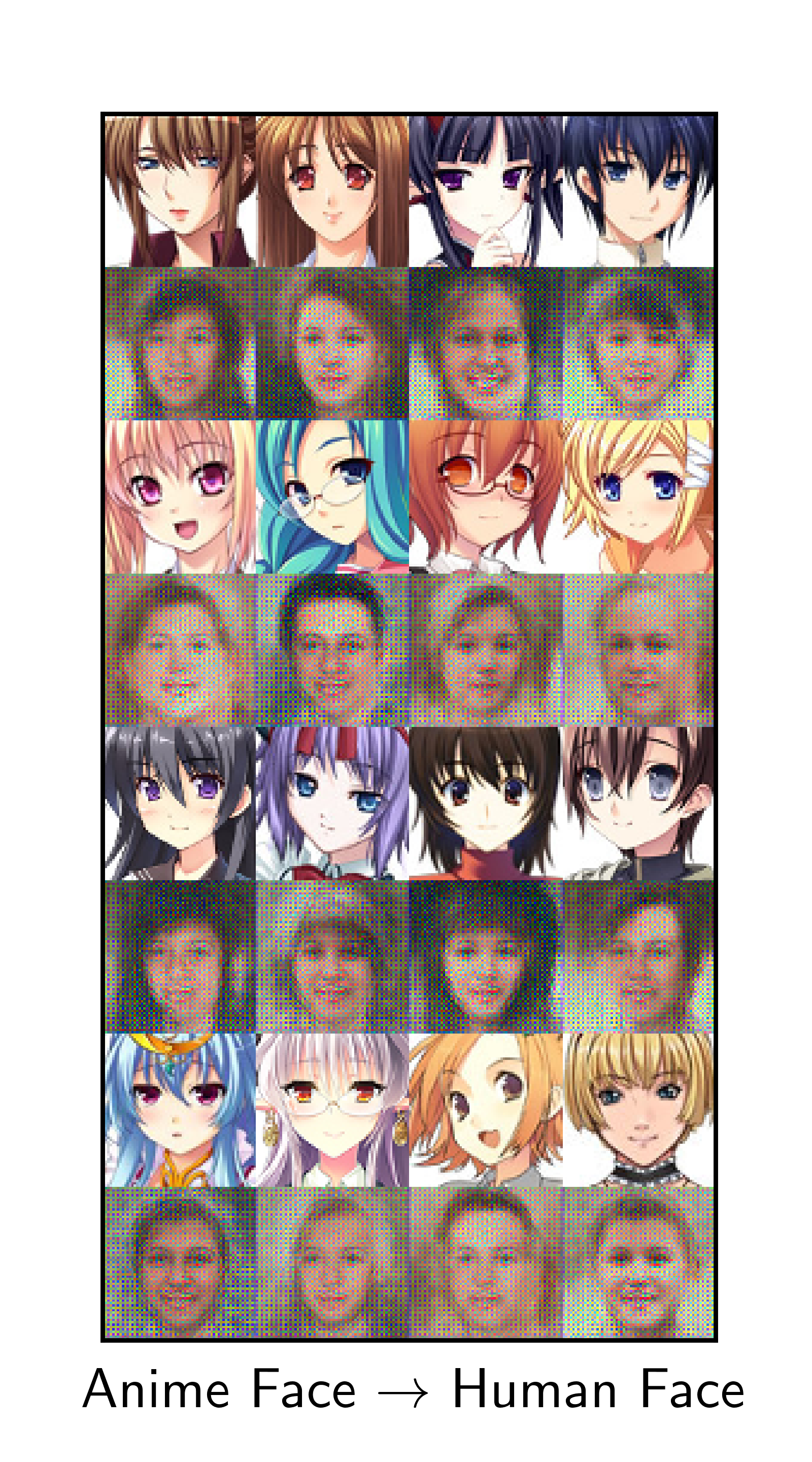}
  \caption{anime faces}
  \label{fig:ba}
\end{subfigure}
\caption{transferring domains of images : \subref{fig:ab} - human face to anime face ;  \subref{fig:ba} - anime face to human face. odd rows : target image ; even rows : converted image}
\label{fig:imtran}
\end{figure}

\begin{figure}[h]
\centering
\begin{subfigure}{0.333\textwidth}
  \centering
  \includegraphics[width =1\linewidth]{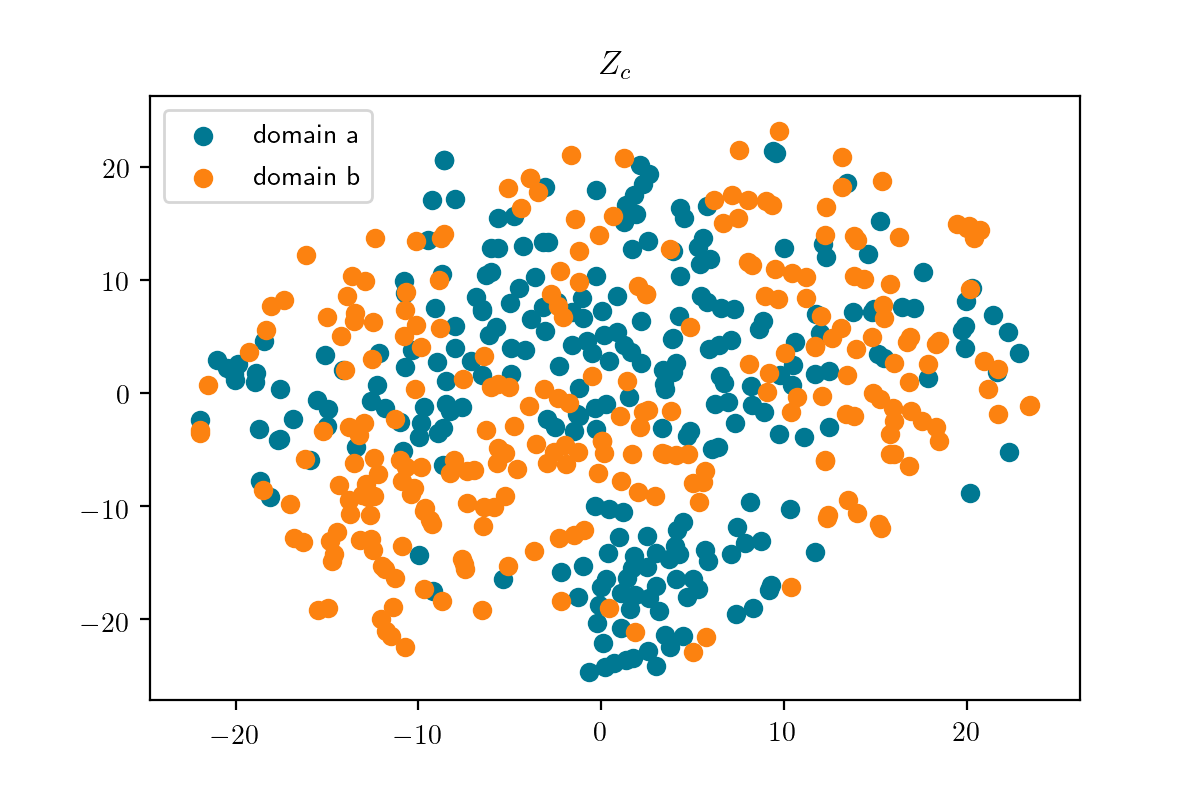}
  \caption{}
  \label{fig:en1}
\end{subfigure}%
\begin{subfigure}{0.333\textwidth}
  \centering
  \includegraphics[width = 1\linewidth]{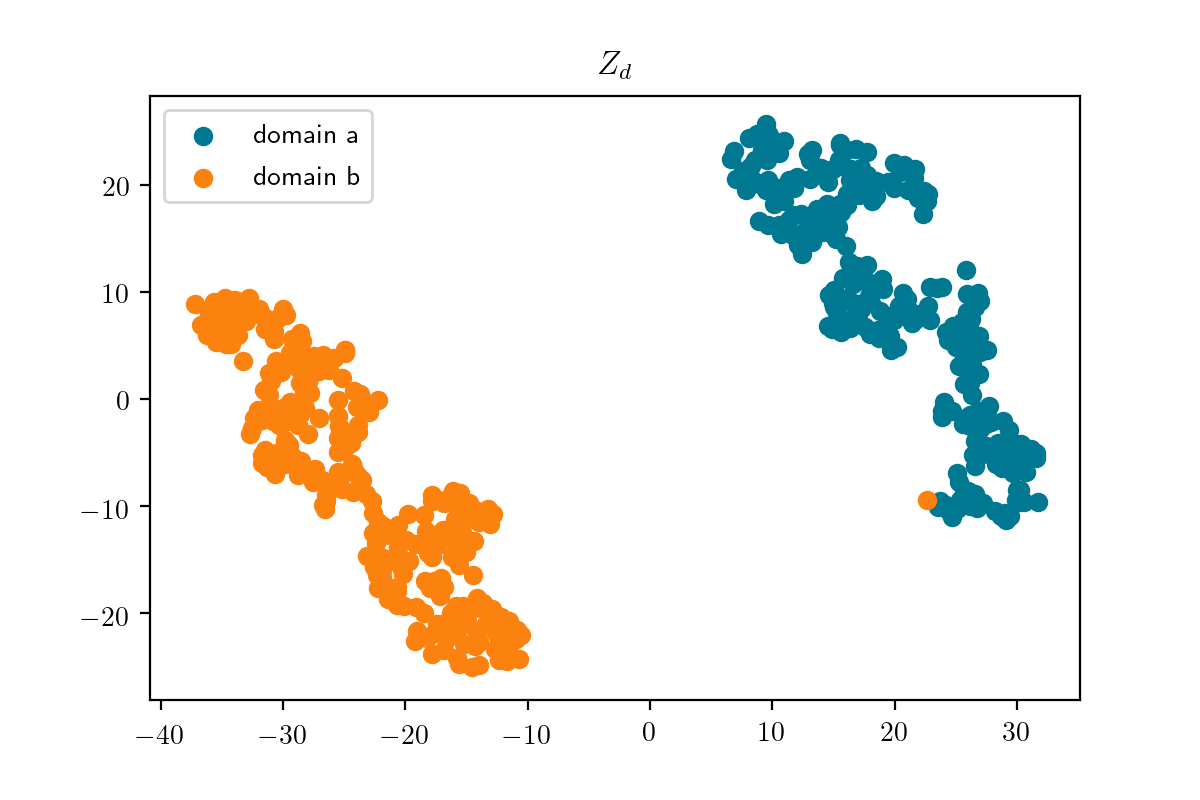}
  \caption{}
  \label{fig:en2}
\end{subfigure}%
\begin{subfigure}{0.333\textwidth}
  \centering
  \includegraphics[width = 1\linewidth]{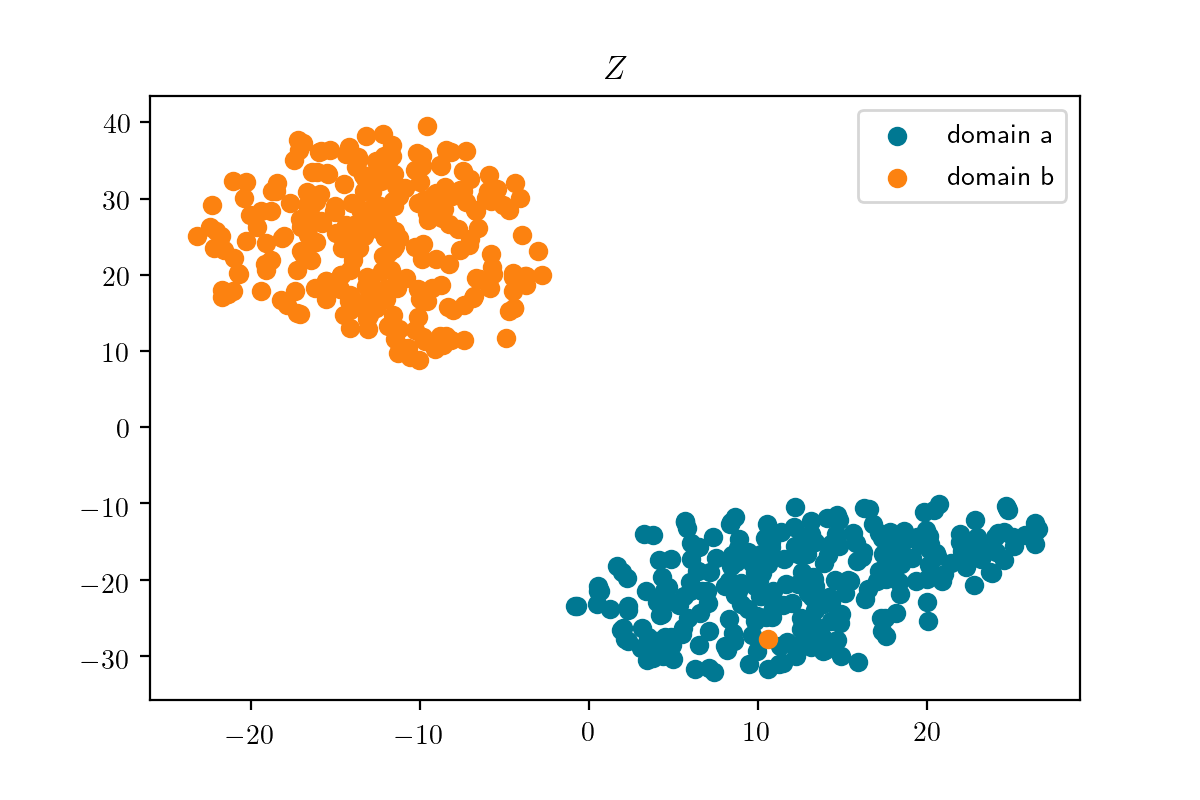}
  \caption{}
  \label{fig:en3}
\end{subfigure}
\caption{t-sne plot of latent encodings of facial images. blue (domain a) : human faces; orange (domain b) : anime fces. (plot \subref{fig:en1}) content encoding ; (plot \subref{fig:en2}) domain encoding ; (plot \subref{fig:en3}) combined encoding}
\label{fig:enc}
\end{figure}

\subsection{Cross-Domain Nearest Neighbours Search}
\begin{figure}[h]
\centering
\includegraphics[width =1\linewidth]{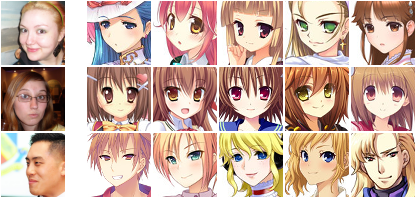}
\caption{left : target image from domain a (human faces) ; right : 5 images from domain b (anime faces)  with the nearest content encoding (leftmost being the closest and rightmost being the farthest) }
\label{fig:nn1}
\end{figure}

Given a target image from the domain $X_1$, we compare it's latent \textbf{content encoding} (figure \ref{fig:nn1}) $Z_c$ to the images from domain $X_2$ and get the images with the closest \textbf{content encoding}. 

\subsection{Image Classification}
For this task, we trained our network (the variation of SRAE with two discriminators) on a dataset consisting of X-Ray Images\citep{xray} of pneumonia patients, scaled down to $64 \times 64$ pixels. We then trained a classifier to predict the domain (a : pneumonia patient ; b : normal) from the \textbf{domain encodings} $Z_d$ and achieved a training accuracy of 84.73\% and a test accuracy of 84.57\% 

\begin{figure}[h]
\centering
\begin{subfigure}{0.333\textwidth}
  \centering
  \includegraphics[width =1\linewidth]{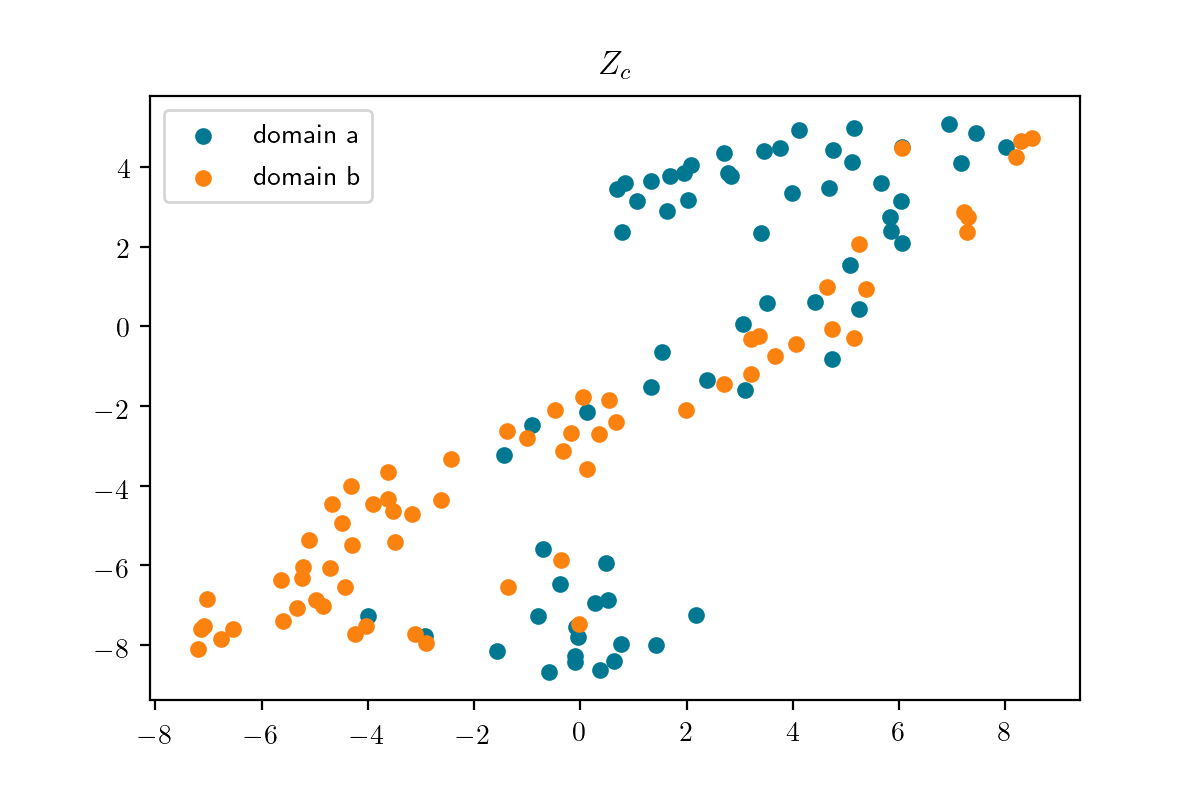}
  \caption{}
  \label{fig:en_p1}
\end{subfigure}%
\begin{subfigure}{0.333\textwidth}
  \centering
  \includegraphics[width = 1\linewidth]{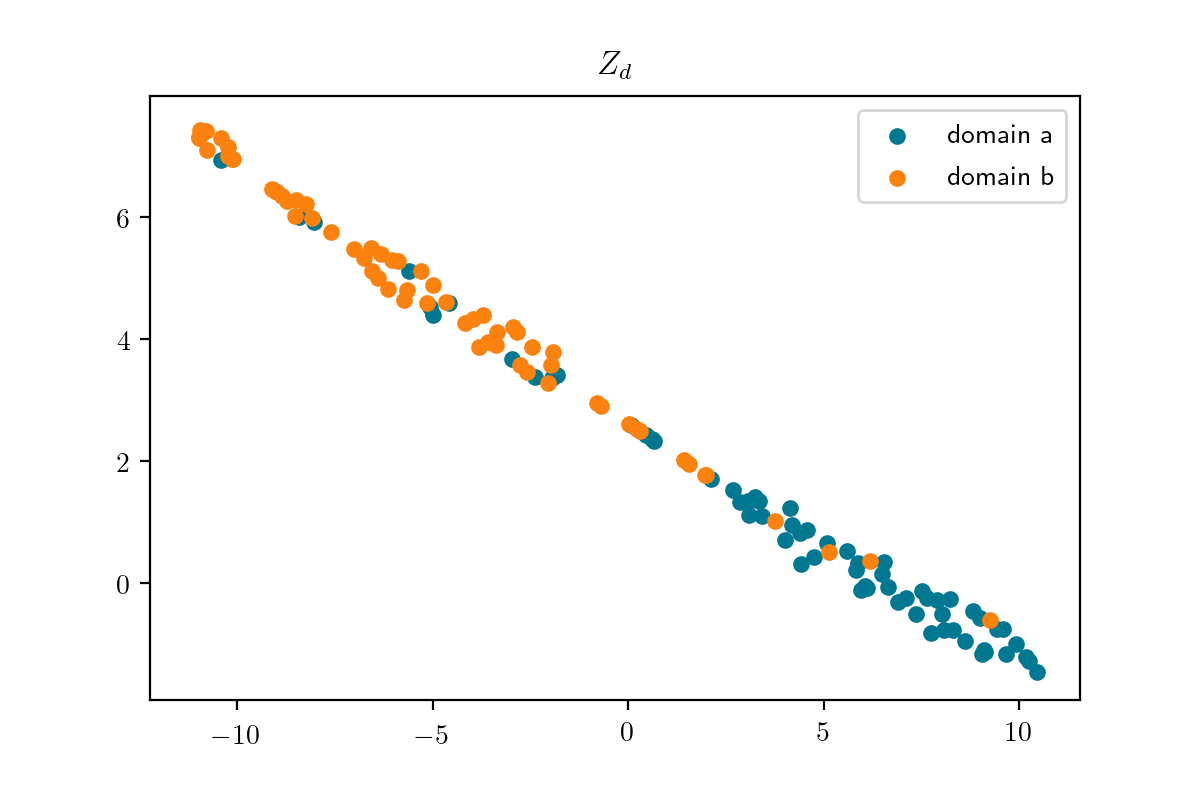}
  \caption{}
  \label{fig:en_p2}
\end{subfigure}%
\begin{subfigure}{0.333\textwidth}
  \centering
  \includegraphics[width = 1\linewidth]{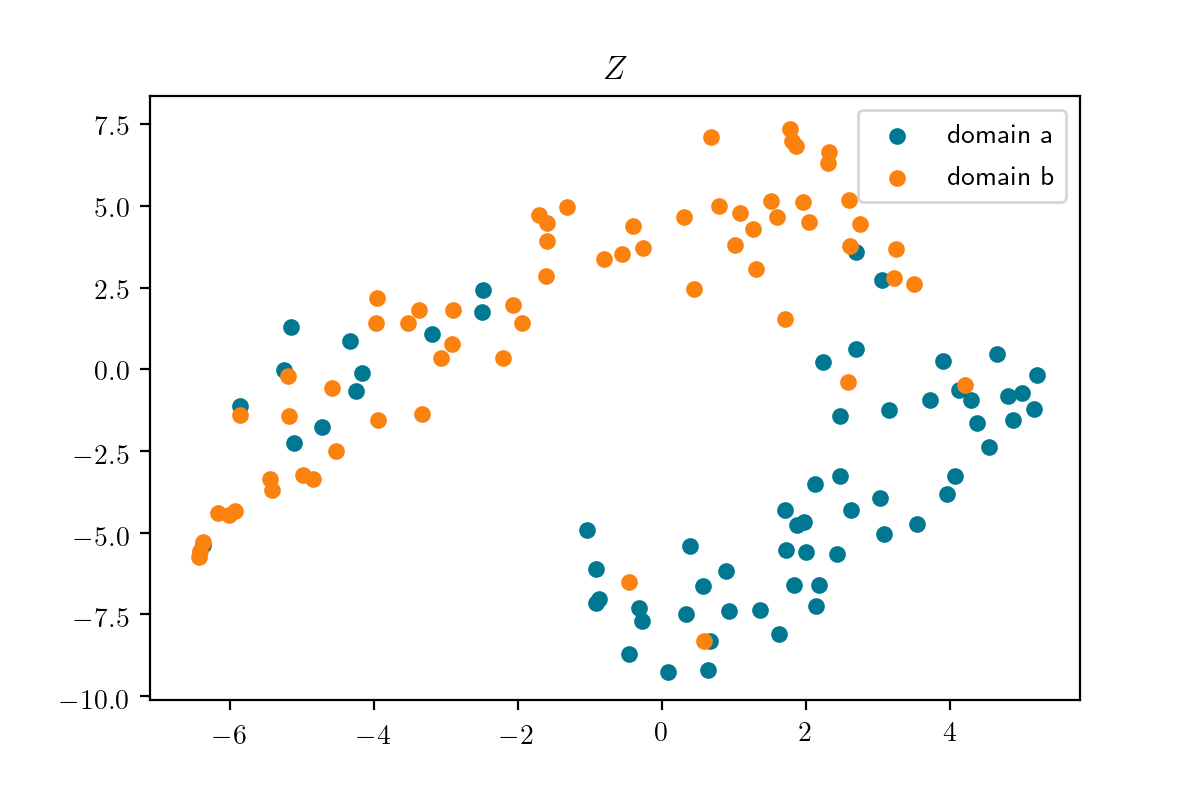}
  \caption{}
  \label{fig:en_p3}
\end{subfigure}
\caption{t-sne plot of latent encodings of x-ray images. blue (domain a) : normal x-ray; orange (domain b) : x-ray of pneumonia patients. (plot \subref{fig:en_p1}) content encoding ; (plot \subref{fig:en_p2}) domain encoding ; (plot \subref{fig:en_p3}) combined encoding}
\label{fig:clas}
\end{figure}

\section{Conclusion and Future Work}
Our current method has a lot of room for improvement. Currently our architecture fails to learn to separate out the latent representations for diverse datasets, CIFAR-10 for example. We would also be explore more into improving the feature consistency during image-to-image translation using our method.

In our upcoming works, we also want to use this architecture for zero-shot learning and making reinforcement learning agents learn to generalize across multiple video-game environments. 

\bibliographystyle{abbrv}
\bibliography{nips20}

\end{document}